\documentclass[conference]{IEEEtran}
\usepackage{cite}
\usepackage{amsmath,amssymb,amsfonts}
\usepackage{algorithmic}
\usepackage{graphicx}
\usepackage{textcomp}
\usepackage{xcolor}
\usepackage{multirow}
\usepackage{tabu}

\usepackage{hyperref}

\hypersetup{
    colorlinks=true,
    linkcolor=blue,
    citecolor=blue,
    filecolor=magenta,      
    urlcolor=cyan,
    pdftitle={Overleaf Example},
    pdfpagemode=FullScreen,
    }
    
\begin{document}

\title{Generative Diffusion Models for Radio Wireless Channel Modelling and Sampling
\thanks{Identify applicable funding agency here. If none, delete this.}
}

\author{\IEEEauthorblockN{1\textsuperscript{st} Ushnish Sengupta}
\IEEEauthorblockA{\textit{MediaTek research} \\
ushnish.sengupta@mtkresearch.com}
\and
\IEEEauthorblockN{2\textsuperscript{nd} Chinkuo Jao}
\IEEEauthorblockA{\textit{MediaTek} \\
chinkuo.jao@mediatek.com}
\and
\IEEEauthorblockN{3\textsuperscript{rd} Alberto Bernacchia}
\IEEEauthorblockA{\textit{MediaTek Research} \\
alberto.bernacchia@mtkresearch.com}
\and
\IEEEauthorblockN{4\textsuperscript{th} Sattar Vakili}
\IEEEauthorblockA{\textit{MediaTek Research} \\
sattar.vakili@mtkresearch.com}
\and
\IEEEauthorblockN{5\textsuperscript{th} Da-shan Shiu}
\IEEEauthorblockA{\textit{MediaTek Research} \\
ds.shiu@mtkresearch.com}
}

\maketitle

\begin{abstract}
Channel modelling is essential to designing modern wireless communication systems. The increasing complexity of channel modelling and the cost of collecting high-quality wireless channel data have become major challenges. In this paper, we propose a diffusion model based channel sampling approach for rapidly synthesizing channel realizations from limited data. We use a diffusion model with a \textit{U Net} based architecture operating in the frequency space domain. To evaluate how well the proposed model reproduces the true distribution of channels in the training dataset, two evaluation metrics are used: \emph{i)} the approximate $2$-Wasserstein distance between real and generated distributions of the normalized power spectrum in the antenna and frequency domains and \emph{ii)} precision and recall metric for distributions. We show that, compared to existing GAN based approaches which suffer from mode collapse and unstable training, our diffusion based approach trains stably and generates diverse and high-fidelity samples from the true channel distribution. We also show that we can pretrain the model on a simulated urban macro-cellular channel dataset and fine-tune it on a smaller, out-of-distribution urban micro-cellular dataset, therefore showing that it is feasible to model real world channels using limited data with this approach.
\end{abstract}


\section{Introduction}

Modeling the wireless channel is an essential step in designing and evaluating the performance of communication systems. In the context of 6G, the wireless environment has become increasingly complex, and the extension of frequency bands and the introduction of large-scale multiple-input multiple-output (MIMO) systems have made it challenging for current channel modeling schemes to accurately reproduce the characteristics of real-world radio channels. Therefore, there is a need for more realistic channel modeling approaches. In addition to accurate channel modeling, the development of deep learning-based wireless communication systems requires large amounts of high-quality wireless channel data to support the training of neural networks (NN). However, the collection of wireless channel data is quite costly and time-consuming, which motivates the development of novel channel data generating solutions to support neural network training.

Previous papers have proposed using machine learning techniques as an alternative to traditional deterministic or stochastic channel models. Markov models have long been used to model the time-varying behaviour of channels \cite{arauz2003markov}. Zander \cite{zander2021applying} showed that a variational autoencoder can be used to generate coefficients of the channel matrix. In Smith \emph{et al.}~\cite{8904907}, a generative adversarial network (GAN) based channel modeling strategy was introduced and it was shown that the GAN can reproduce a simple additive white Gaussian noise channel. Xiao \emph{et al.}~\cite{xiao2022channelgan} proposed ChannelGAN, a Wasserstein GAN (WGAN) that can generate the multiple-input multiple-output (MIMO) channel matrix. The MIMO-GAN approach \cite{orekondy2022mimo} also uses a Wasserstein GAN to model the distribution of channels, however unlike the ChannelGAN paper, they use pairs of input-output signals as their data and the channel impulse response is modeled implicitly by the neural network.

Despite multiple studies that have explored GANs as a way of modeling the channel from data, they have several significant drawbacks that have been highlighted in recent research. One of the most significant challenges with GANs, identified first in the original GAN paper itself \cite{goodfellow2020generative}, is mode collapse, where the generator learns to generate only a limited set of samples, ignoring the rest of the data distribution. This results in a lack of diversity in the generated samples, which can be a major problem for a channel simulator. GANs are also notoriously difficult to train, with many models failing to converge or producing low-quality samples. This instability is due to the adversarial nature of the training process, where the generator and discriminator are constantly trying to outwit each other. The Wasserstein GAN used in the ChannelGAN paper \cite{xiao2022channelgan} employs a different loss function with a gradient penalty to address this issue and has been shown to be more stable in practice, but does not resolve it entirely. 

Denoising diffusion probabilistic models are a class of generative models that have recently gained popularity due to their ability to generate high-quality images, video and audio. Diffusion models are based on the principle of diffusing noise through a sequence of invertible transformations, where the noise is initialized as a standard Gaussian distribution and is gradually transformed into the target data distribution. The process can be reversed to generate new samples from the target distribution. One advantage of diffusion models over GANs is that they do not require a discriminator network to guide the generator network during training. Instead, diffusion models directly optimize the likelihood of the target data distribution. This is why their training is also much more stable. Ho \emph{et al.} \cite{ho2020denoising} compared diffusion models and GANs on several image generation tasks and found that diffusion models outperformed GANs in terms of image quality and diversity, especially for high-resolution images. The authors attributed the superior performance of diffusion models to their ability to capture the full distribution of the data, while GANs only learn a single mode or a subset of the modes of the distribution. In this paper, we propose using diffusion models to learn samples from the distribution of the channel impulse response matrices from data. 

The existing literature on this topic also does not focus too much on measuring how much the generated channels overlap with the distribution of the original data. To compare the distribution of generated channel matrices with the real ones, we first compute the approximate Wasserstein distance between the distributions of real and generated channel power spectra. We also consider precision and recall as a measure to better separately capture the fidelity and diversity of the generated images. 

Simulations can generate copious amounts of channel data but the datasets collected from real-world experimental campaigns are limited in size. This is a problem for data-hungry machine learning approaches to channel modelling. Our proposed solution involves transfer learning: we show that we can pre-train the diffusion model on a simulated dataset and fine-tune it on a smaller, out-of-distribution dataset, therefore showing that it is feasible to model real world channels using limited data with this approach.

\section{Methods}

\subsection{Comparison with Wasserstein GAN }

\begin{figure}[t]
\centering
\includegraphics[width=9cm]{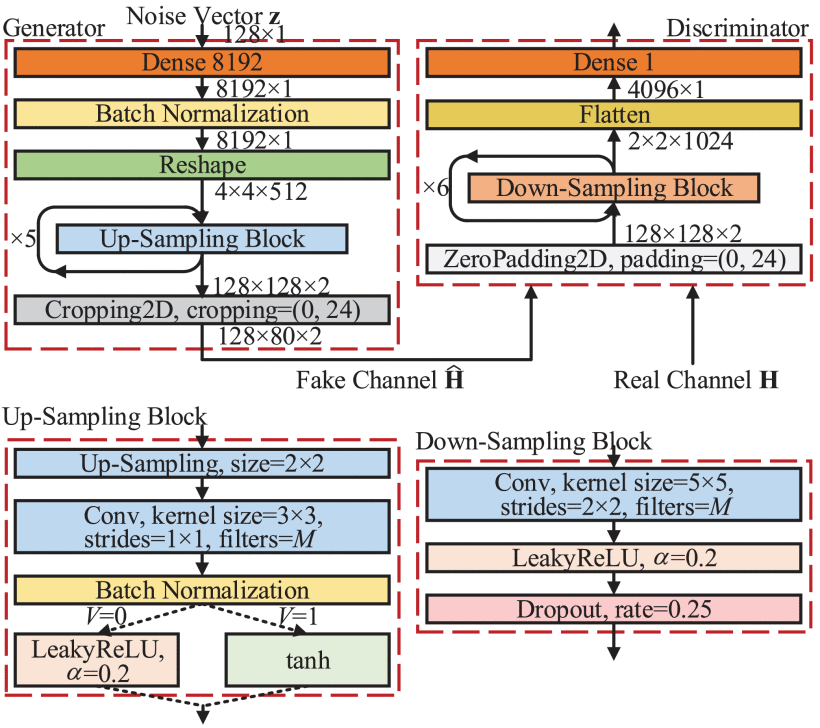}
\caption{The ChannelGAN architecture \cite{xiao2022channelgan}}
\end{figure}

As discussed in the introduction, WGANs have so far shown promising results in being able to model the distribution characteristics of real channel matrices, so we compare our diffusion models with a WGAN. We borrow the ChannelGAN architecture \cite{xiao2022channelgan} with minor modifications. For the generator $G$, it maps a noise vector $z$ to a generated channel $H \in \mathbb{R}^{N_a \times N_{f'} \times 2}$, where $N_a = N_t \times N_r$ denotes the number of antenna pairs, $N_{f'}$ denotes the number of frequency bins retained (the high frequency portion of the response is cropped off). Firstly, in order to convert from input noise vector to the feature maps, the noise vector $z$ is processed by a dense layer along with a reshape. Next, five up-sampling blocks are stacked up. Specifically, each up-sampling block is composed of a $2 \times 2$ up-sampling layer (nearest neighbor interpolation), a $3 \times 3$ convolutional layer (Conv) with $M$ filters and a batch normalization layer. Two optional activation functions are also attached at the tail of each block with the opting flag $V \in {0, 1}$, where $V = 0$ and $V = 1$ indicate utilization of leaky rectified linear unit (LeakyReLU) and hyperbolic tangent function (tanh), respectively. Here five up-sampling blocks are configured with parameters $M = \{1024, 512, 256, 128, 2\}$ and $V = \{0, 0, 0, 0, 1\}$ sequentially. Note that the tanh in the last Up-Sampling Block limits the amplitude of the elements in fake channel in $[-1, 1]$ which matches the real channel, and the LeakyReLU in remaining blocks brings the non-linear transformation to the network. Finally, the two-dimensional cropping layer (Cropping2D) is used to maintain the shape of generated fake channel $H$ consistent with the real channel $H$.

For the discriminator $D$, the input is firstly padded by a two-dimensional zero padding layer (ZeroPadding2D). Then, six down-sampling blocks are stacked up, where the $5 \times 5$ Conv with the strides of $2 \times 2$ and $M = \{32, 64, 128, 256, 512, 1024\}$ filters, a LeakyReLU and a dropout layer are deployed sequentially. In the end, a flatten operation and dense layer with single output are utilized.

\subsection{Diffusion models}

Denoising diffusion probabilistic models (DDPMs) are a new class of generative models introduced by Sohl-Dickstein and colleagues \cite{sohl2015deep} following a line of research on Markov-chain based generative models \cite{bengio2014deep, salimans2015markov}. These methods are based on iteratively corrupting the data to transform the data distribution into a known simple distribution, often an isotropic normal distribution, and learning to reverse the process by estimating the mean for the reverse iterations using a neural network. References \cite{song2019generative, song2020improved} proposed a similar algorithm that learns the score of the data distribution in the forward process and uses the Langevin dynamic to reverse the process and sample from the distribution. 

The diffusion model generates a sequence of samples by iteratively adding noise to the previous samples. At each step, the forward diffusion process adds Gaussian noise with a variance determined by a noising schedule to the previous sample. Given a data-point $\textbf{x}_0$ sampled from the real data distribution $q$, one can define a forward diffusion process by adding noise stepwise. The forward diffusion process is performed for $T$ steps:
$$\textbf{x}_{t+1} = \sqrt{1-\beta_t}\textbf{x}_t + \sqrt{\beta_t}\epsilon_t$$

where $\textbf{x}_t$ is the noisy datapoint at step $t$, $\textbf{x}_{t+1}$ is the noisy datapoint at step $t+1$, $\beta_t$ is the noise level at step $t$ which is determined according to a cosine noising schedule \cite{nichol2021improved} and $\epsilon_t$ is a Gaussian noise with zero mean and unit variance.

As $T$ becomes large, the latent $\textbf{x}_T$ approaches an isotropic Gaussian distribution. Therefore, if we manage to learn the reverse distribution $q(\textbf{x}_{t-1}|\textbf{x}_t)$ using a neural network, we can sample $\textbf{x}_T$ from the unit normal distribution, run the reverse process and acquire a sample from $q(\textbf{x}_0)$, generating novel data points from the original data distribution.

\begin{figure}[t]
\centering
\includegraphics[width=8cm]{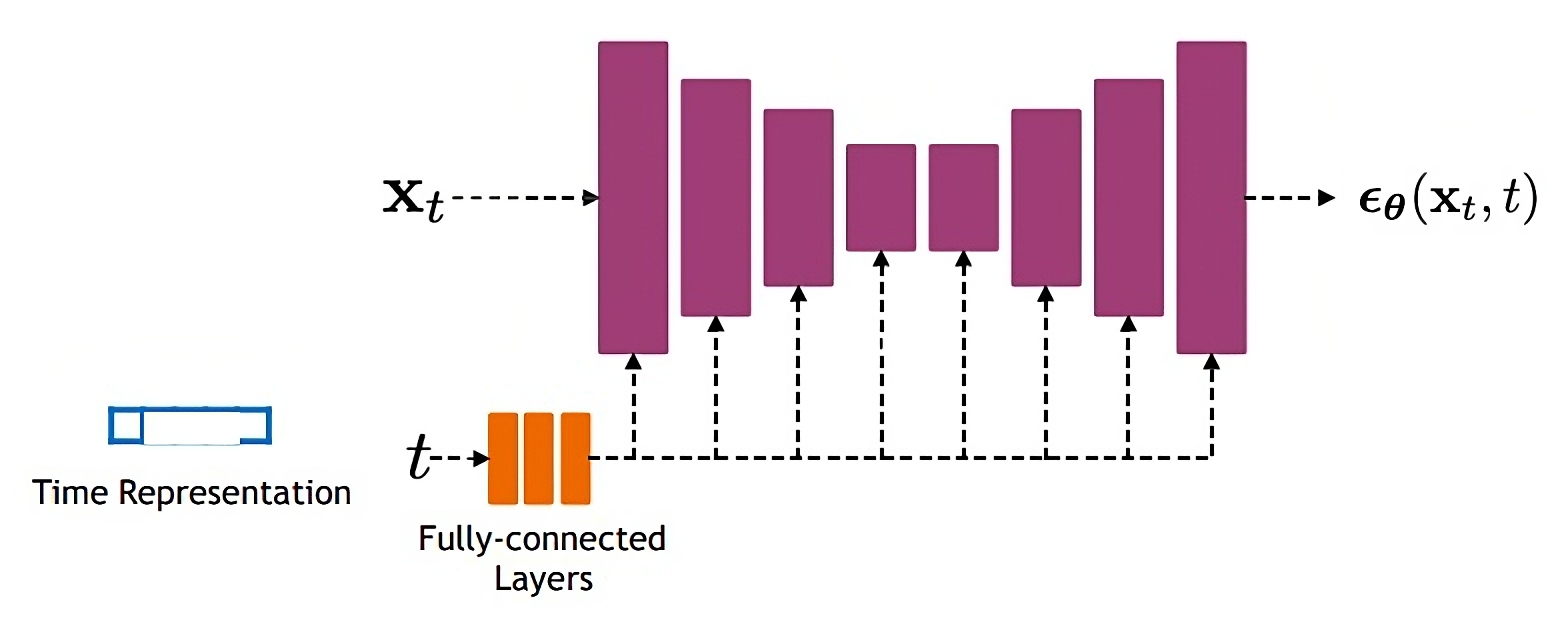}
\caption{The U-Net architecture used for denoising in our diffusion models}
\end{figure}

A neural network with a U-Net architecture (Figure 1) is used to predict the added noise $\epsilon_t$ given a noisy input. The U-Net is trained to minimize the mean squared error between the predicted noise and the true noise. The architecture consists of a contracting path and an expanding path. The contracting path consists of a series of convolutional layers with max-pooling. The expanding path consists of a series of convolutional layers with up-sampling. The contracting and expanding paths are connected by skip connections. The skip connections enable the network to learn both low-level and high-level features.

\subsection{Fine-tuning the pretrained model}
Few-shot generation seeks to generate more data of a given domain, with only few available training examples. As it is unreasonable to expect to fully infer the distribution of real channels from just a few observations, we seek to leverage a large, related source domain (simulated urban macro data) for pre-training. Thus, we wish to preserve the diversity of the source domain, while adapting to the appearance of the target. We adapt our pre-trained model, without introducing any additional parameters, to the fewer examples of the target domain, by training on the new data using a lower learning rate. Crucially, we regularize the changes of the weights during this adaptation, in order to best preserve the information of the source dataset, while fitting the target. We demonstrate the effectiveness of this approach by generating high-quality samples from the urban microcellular scenario.

\subsection{Performance metrics}

For generative models, it is important to quantify both the quality and the fidelity of the generated data individually. Let $P_g$ denote the generated channels and $P_r$ denote the test dataset. We compute the normalized power spectra of the channels in the frequency domain and the antenna domain representations. We fit a multivariate Gaussian to the generated and real power spectra. Let ($\mu_{g}$, $\Sigma_{g}$) and ($\mu_{r}$, $\Sigma_{r}$) denote the mean and covariances of the two Gaussians. The metric we will use to measure how close the real and generated distributions are will be the Wasserstein-2 distance between these two distributions:
\begin{equation}
    W_2(P_g,P_r) = ||\mu_g - \mu_r||^2 + \text{Tr}(\Sigma_g+\Sigma_r-2(\Sigma_r^{1/2}\Sigma_g \Sigma_r^{1/2})^{1/2})
\end{equation}

\begin{figure}[t]
\centering
\includegraphics[width=9cm]{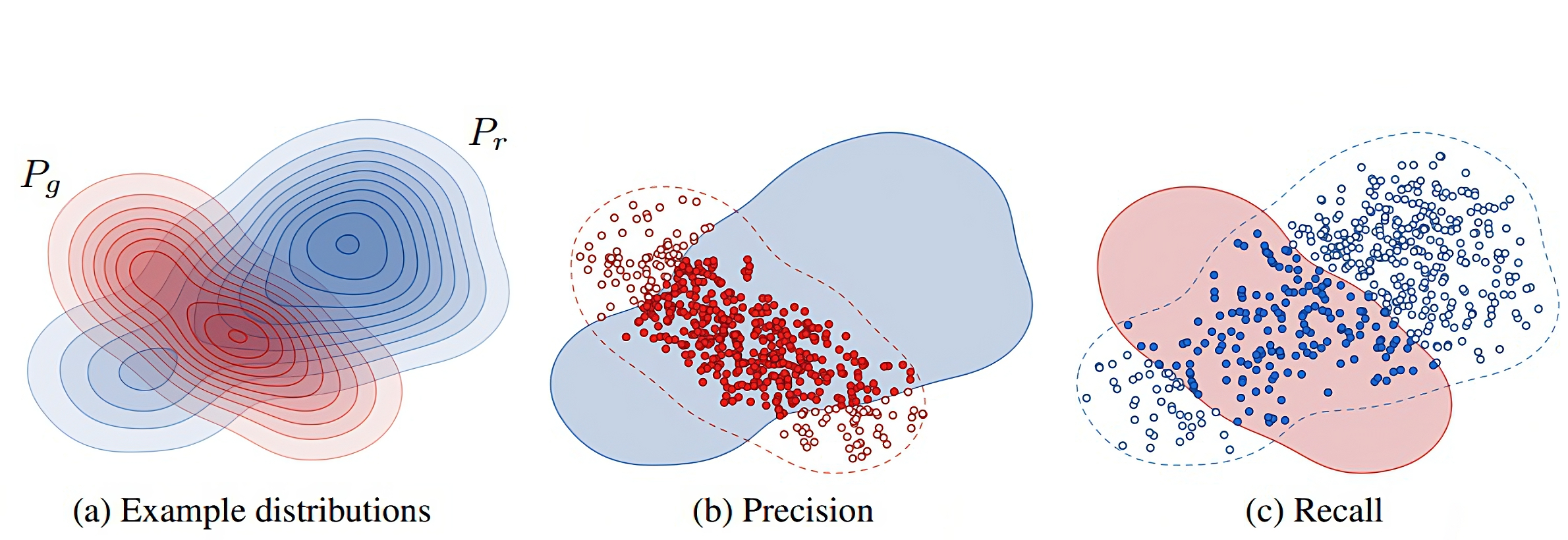}
\caption{Diagram illustrating precision and recall for distributions \cite{kynkaanniemi2019improved} We denote the distribution of real data with $P_r$ (blue) and the distribution of generated data with $P_g$ (red). Precision is the probability that a random sample from $P_g$ falls within the support of $P_r$. Recall is the probability
that a random sample from $P_r$ falls within the support of $P_g$.}
\end{figure}

However, this approximate Wasserstein distance isn't the best measure, because it uses a Gaussian approximation to a non-Gaussian distribution and it also ignores the phase of the complex channel response. It also fails to disentangle the accuracy and variety of generated channels. A metric that better captures the fidelity and diversity of the generated channels is the improved precision and recall metric proposed by Kynk\"a\"anniemi \emph{et al.} \cite{kynkaanniemi2019improved}. The key idea is to draw an equal number of samples from real and generated distributions and embed them into a lower dimensional feature space using a pre-trained autoencoder network. We then calculate the pairwise Euclidean distances between all feature vectors in the set and, for each feature vector, form a hypersphere with radius equal to the distance to its nearest neighbor. This hypersphere defines a volume in the feature space that serves as an estimate of the true manifold. To determine precision, we query for each generated channel whether it is within the estimated manifold of real channels. For recall, we query for each real channel whether it is within estimated manifold of generated channels. 

\section{Dataset description}

The channel dataset is generated from a system-level simulator (SLS) with a 3D stochastic channel model. The channel generation procedure follows the steps described in TR 38.901 \cite{zhu20213gpp} A total of 120 User Equipments (UEs) are dropped in a hexagonal network layout and associated with the serving Base Station (BS) with maximum reference signal received power (RSRP). For each UE-BS link, the small-scale parameters of the channel angle (AoD, AoA, ZoD, ZoA) and delay information are created independently. Based on these parameters, channel impulse responses are generated depending on BS and UE antenna settings. We also considered the spatial consistency procedure defined in TR 38.901. Therefore, the channel data of adjacent UEs are correlated. Two scenarios UMa and UMi channels are used for different training purposes. 32000 samples of channel impulse response matrices from 96 UEs in each scenario are used as training data and 8000 channel matrices from the remaining 24 UEs are used for testing and validation.

\begin{table}
    \centering
    \caption{Channel statistics and parameters setting}
    \begin{tabular}{|c|c|c|c|c|}
   \hline
   \hline
         \multirow{2}{4em}{Scenario} & \multicolumn{2}{|c|}{Urban Macro (UMa)} & \multicolumn{2}{|c|}{Urban Micro (UMi)}\\
    \cline{2-5}
         & LOS & NLOS & LOS & NLOS \\
         \hline
         Number of clusters&$12$&$20$&$12$&$19$ \\
         \hline
         Delay spread (ns) &$103.8$&$455.6$&$55.7$&$113.6$ \\
         \hline
         AoD spread (degree) &$12.6$&$29.2$&$15.4$&$26.3$ \\
         \hline
         AoA spread (degree) &$64.6$&$99.7$&$49.2$&$59.1$ \\
         \hline
         Inter-site distance (ISD) &\multicolumn{2}{|c|}{$500 m$}&\multicolumn{2}{|c|}{$200 m$}\\
         \hline
         BS antenna height &\multicolumn{2}{|c|}{$25 m$}&\multicolumn{2}{|c|}{$10 m$}\\
         \hline
         Career frequency & \multicolumn{4}{|c|}{$2$ \emph{GHz}}\\
         \hline
         UE speed & \multicolumn{4}{|c|}{$3$ and $30$ \emph{km/hr}}\\
         \hline
         BS antenna setting & \multicolumn{4}{|c|}{$(M,N,P, M_{p}, N_{p}) =(4,4,2,1,4)$}\\
         \hline
         UE antenna setting & \multicolumn{4}{|c|}{$(M,N,P, M_{p}, N_{p}) =(1,2,2,1,2)$}\\
         \hline
         \hline
    \end{tabular}
    \label{tab:data}
\end{table}

To demonstrate that our model can be pre-trained on a corpus of simulated data and fine-tuned on a limited real-world dataset, we fine tune the diffusion model trained on urban macro-cell scenarios on a dataset of 20000 (different proportions of the dataset are used in the fine-tuning experiments) channel impulse responses from a urban micro-cell simulation. Compared to the simulated urban macro channels, the urban micro channels have a different distribution of arrival/ departure angles as well as delay spreads, and acts as a stand-in for real data.

As shown in Table 1, the inter-site distance (ISD) and BS height of UMa are greater than those of UMi. With wider coverage, UMa is a richly scattered environment with larger channel delay and angle spread characteristics. Both LOS and NLOS conditions are considered in our datasets. At the BS side, we consider a 2D antenna array, $(M,N,P,M_{p},N_{p})=(4,4,2,1,4)$, where $N$ is the number of columns, $M$ is the number of antenna elements with the same polarization in each column. $M_{p}$ and $N_{p}$ is the number of antenna ports in row and column. $P=2$ means that polarization antennas are used. Therefore, four vertical antenna elements are virtualized to one antenna port in each column. A total of eight antenna ports are used in the channel model.

The MIMO channel impulse response in the frequency domain can be denoted as a complex tensor $\mathbf{H} \in \mathbb{C}^{N_t \times N_r \times N_f}$, where $N_t$ is the number of transmitting antennae, $N_r$ is the number of receiving antennae and $N_f$ represents the number of frequency bins. For training our generative models, we reshape and crop the tensor so that it becomes a real-valued tensor with two channels $\in \mathbb{R}^{N_a \times N_{f'} \times 2}$, where $N_a = N_t \times N_r$ denotes the number of antenna pairs, $N_{f'}$ denotes the number of frequency bins retained (the high frequency portion of the response is cropped off) and the last dimension with a value of 2 represents the real and imaginary parts of original complex channel, respectively. Our objective is to generate samples of channel matrices $\mathbf{H} \sim q(\mathbf{H})$ from the true distribution of $\mathbf{H}$, given a limited number of channel samples as our training data.

\section{Results}
\subsection{Performance and comparison with WGAN}

\begin{figure*}[t]
\centering
\includegraphics[width=18cm]{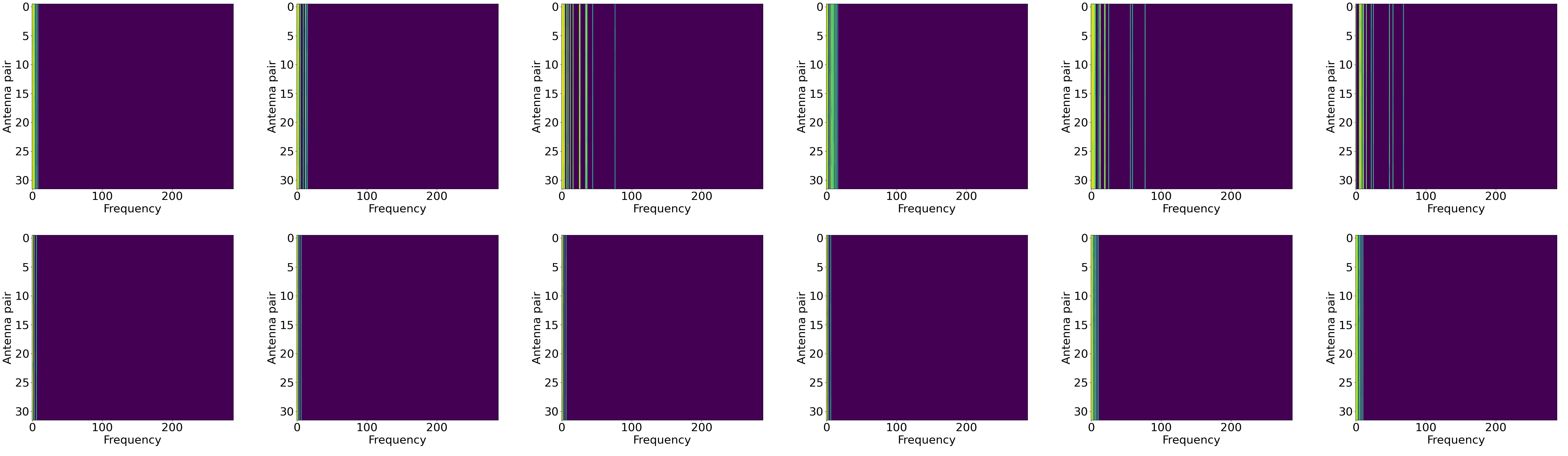}
\caption{Normalized power spectra of urban macrocellular channel impulse response samples generated by the diffusion model (top row) and the GAN (bottom row). The diffusion model clearly generates more diverse channel samples and this is confirmed by our metrics.}
\end{figure*}

In this section, we walk through the evaluation of our diffusion model approach and the Wasserstein GAN. Both models were trained and evaluated using the same urban macrocell channel dataset described in Section 3. In figure 4, we compare the training progress of the diffusion model and the Wasserstein GAN by plotting the 2-Wasserstein distance of the antenna-domain power spectra between generated channels and real channels from the test dataset against the number of epochs the model has trained for. As discussed in the methods section, this metric is an approximate measure of how well the distribution of generated channels matches the distribution of real channels. We observe that the diffusion model converges stably to a fairly low Wasserstein distance, but the GAN experiences episodes of periodic instability throughout the training process due to the adversarial nature of the training. As a result, even the GAN with the best set of hyperparameters cannot outperform in terms of 2-Wasserstein distance (refer to Table \ref{tab:results}).

\begin{figure}[t]
\centering
\includegraphics[width=8cm]{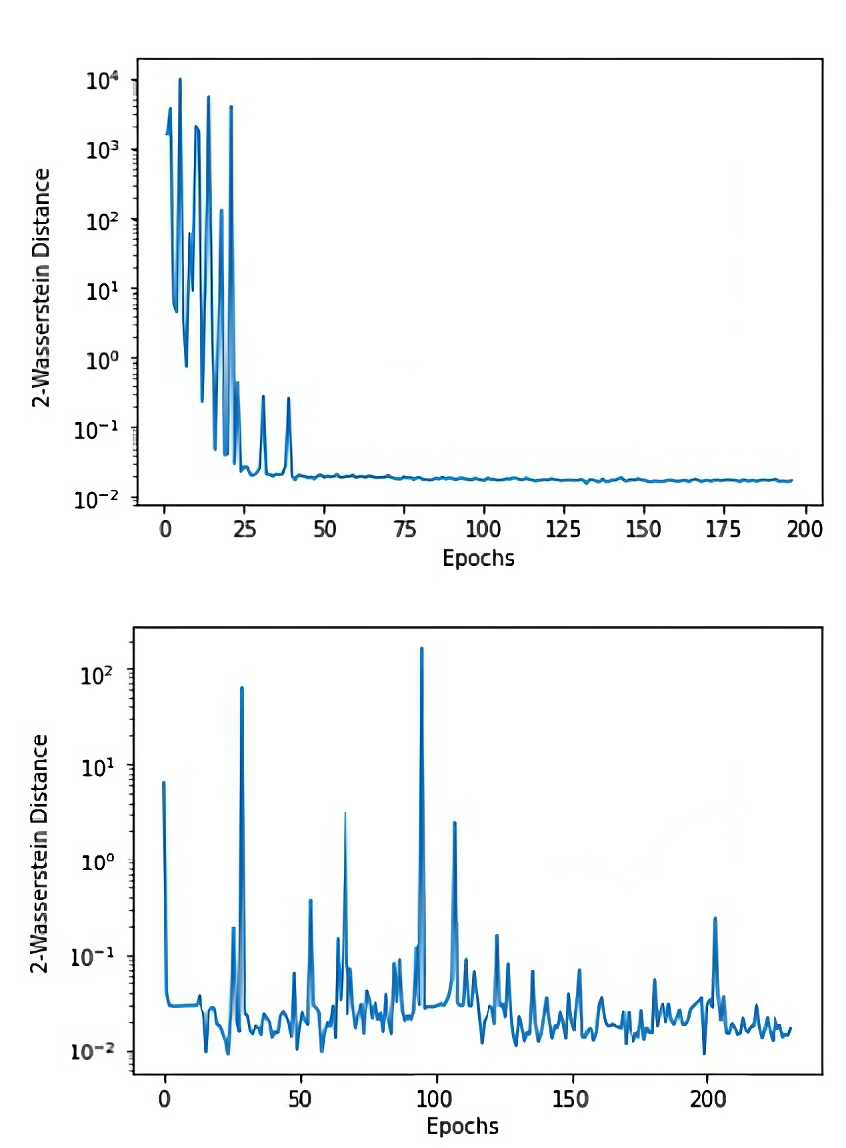}
\caption{The evolution of the 2-Wasserstein distance in antenna domain between the real and generated power spectra in the antenna domain for the diffusion model (top) and the WGAN (bottom) during training. The large spikes in the training curve of the WGAN even in later epochs indicate that it suffers from instabilities throughout the training process.}
\end{figure}

In table \ref{tab:results}, we also compare the performance of the WGAN with the diffusion models in terms of precision and recall, to understand how faithfully each model captures real channels and if the variety of the generated samples match that of the data. Unsurprisingly, we find that the WGAN is fairly competitive with the diffusion model when it comes to precision, which measures the fidelity of generated samples. However, the WGAN is much worse when it comes to recall, which corresponds to the diversity of the generated data. This indicates that the Wasserstein GAN has suffered from at least partial mode collapse. This is also apparent from the channel samples generated by each model (Figure 4); the diffusion model generates channels that are visibly more diverse. 

\begin{table}[t!]
\centering
\caption{Comparison of the Wasserstein distance, precision and recall computed on the validation dataset for the diffusion model and WGAN.}
\begin{tabular}{||c c c c c||} 
 \hline
 \  & $W_2$ (antenna) & $W_2$ (frequency) & Precision & Recall \\ [0.5ex] 
 \hline\hline
 WGAN & 0.0017 & 0.0029 & 0.8350 & 0.5750 \\ 
 DDPM (ours) & 0.0010 & 0.0023 & 0.8210 & 0.8950 \\ [1ex] 
 \hline
\end{tabular}

\label{table:1}
\label{tab:results}
\end{table}

\subsection{Performance of pretrained model after fine-tuning}

For the fine-tuning experiments, we use different proportions of urban microcellular data for fine-tuning ($5\%$, $10\%$, $25\%$ and $100\%$ of the dataset) and compare its performance with a model trained from scratch with the same amount of urban microcellular data. We find that the fine-tuned models achieve considerably better precision and recall, and this effect is particularly visible when using a small fraction ($5\% - 10\%$) of the data. Even when using all of the data from the out-of-distribution, the fine-tuned model maintains a small advantage over the trained-from-scratch problem, which indicates that there is no significant negative transfer for this problem. We also observe that fine-tuning has a more pronounced effect on recall (diversity) than precision (fidelity). The plots of precision and recall in figure 6 summarizes the results of the fine-tuning experiments. These results are very encouraging and indicates that a model pre-trained on simulated channel data can learn to generate channels from a related but different distribution, such as real world channels, using much less data.

\begin{figure}[t]
\centering
\includegraphics[width=8cm]{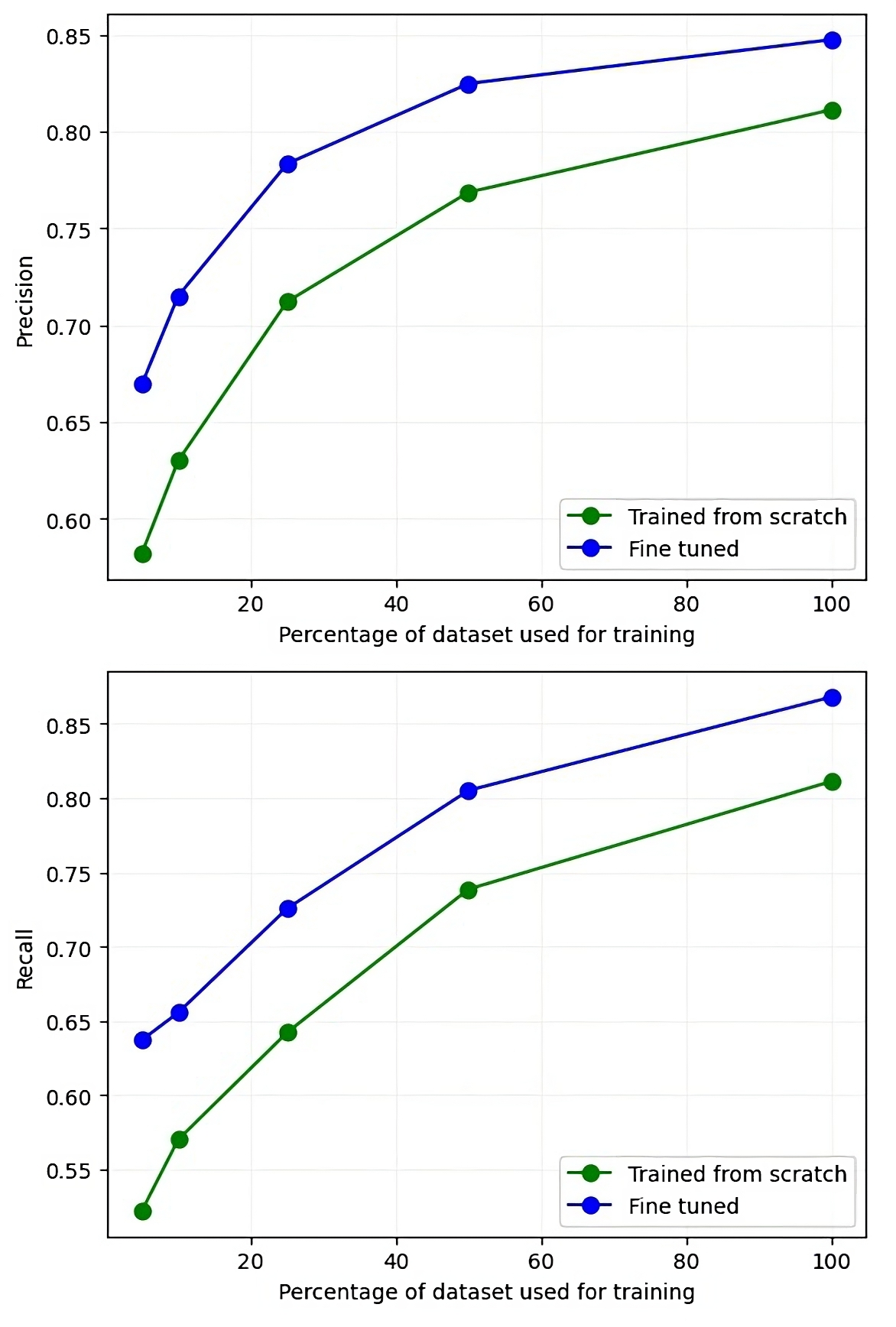}
\caption{Comparison of precision (top) and recall (bottom) metrics between a diffusion model pre-trained on the urban macro dataset/ fine-tuned on urban micro dataset and a model trained from scratch on the urban micro dataset. The fine-tuned model clearly has better precision and recall, especially when a low proportion of the target dataset (urban microcellular) is used for training.}
\end{figure}

\section{Conclusions}

In this paper, a diffusion model based wireless channel modeling framework has been proposed and analyzed. Unlike traditional methods which involve detailed theoretical analysis and data processing to derive key channel parameters from real measurement data, our new method does not require any domain-specific knowledge or technical expertise, and can obtain the target channel model by directly learning from raw MIMO channel data with a diffusion model, which iteratively denoises isotropic Gaussian noise to generates samples from the true data distribution. Using synthetic channel data from a 3D stochastic model as our proving ground, the distribution of generated channel samples has been compared with that of the real channel data using approximate Wasserstein distance between power spectra distributions as well as precision and recall metrics from the generative machine learning literature. These metrics help us measure the fidelity and diversity of the generated channels seperately. We find that the diffusion models not only generate high-fidelity channels, but also manage to capture the diversity of the data very well, unlike the WGAN used in previous studies for channel modelling. This is an important property for a channel emulator to have, since it needs to be able to emulate all possible channel conditions. We also learned that diffusion models are much less temperamental to train compared to the GANs, which suffer from frequent instabilities during training.

Using a simulated out-of-distribution dataset (the Urban Micro scenario) as our stand-in for real channel data, we demonstrated that we can use fine-tuning to help a model pre-trained on simulated data generalize to real world channel data, which have a different distribution. We showed that reaching the same level of precision and recall is possible with the pretrained model using 2-3 times less data compared to training from scratch. This is promising as it is much more expensive to collect real channel data, compared to simulations. 

Our future work will include more experiments with real MIMO channel datasets and validation of our synthetic channel model on downstream tasks such as CSI feedback, position estimation from channel data and channel equalization. Another interesting challenge is modelling the time-variation of channels. Advances in video diffusion models \cite{ho2022video} might hold the key to modelling time-variations in statistical channels.

\bibliographystyle{ieeetr}
\bibliography{references}

\newpage

\end{document}